\documentclass{llncs}
\usepackage{placeins}
\usepackage[T1]{fontenc}
\usepackage{graphicx}
\usepackage{amsmath,amssymb}
\usepackage{mathptmx}
\usepackage{hyperref}
\usepackage{booktabs}
\usepackage{float}
\begin{document}

\title{Feature Selection Empowered BERT for Detection of Hate Speech with Vocabulary Augmentation}

\author{Pritish N. Desai\inst{1}\orcidID{0009-0008-6728-1835} \and
Tanay Kewalramani\inst{1}\orcidID{0009-0000-9271-3427}
Srimanta Mandal\inst{1}\orcidID{0000-0003-3871-6621} 
\authorrunning{Desai et al.}
\institute{Dhirubhai Ambani University (Formerly DA-IICT), Gandhinagar, India \\
\email{\{202201312, 202201362, srimanta\_mandal\}@daiict.ac.in}}
}
\maketitle
\begin{abstract}
Abusive speech on social media poses a persistent and evolving challenge, driven by the continuous emergence of novel slang and obfuscated terms designed to circumvent detection systems. In this work, we present a data-efficient strategy for fine-tuning BERT on hate speech classification by significantly reducing training set size without compromising performance. Our approach employs a TF–IDF-based sample selection mechanism to retain only the most informative 75\% of examples, thereby minimizing training overhead. To address the limitations of BERT’s native vocabulary in capturing evolving hate speech terminology, we augment the tokenizer with domain-specific slang and lexical variants commonly found in abusive contexts. Experimental results on a widely used hate speech dataset demonstrate that our method achieves competitive performance while improving computational efficiency, highlighting its potential for scalable and adaptive abusive content moderation. The source code for this work is available at: \url{https://github.com/UltimateTanA/BMP}.
\end{abstract}

\section{Introduction}

The detection of hate speech on social media remains a pressing challenge due to the dynamic and adversarial nature of online language~\cite{ali2025}. Users continuously adapt their expressions by employing creative spellings, deliberate obfuscations, and evolving slang to evade automatic moderation systems. As a result, static or overly generic detection models quickly become outdated, failing to recognize emerging abusive patterns.

Pre-trained language models such as BERT have demonstrated impressive results in various natural language processing (NLP) tasks, including hate speech detection~\cite{caselli-etal-2021-hatebert,Wang2021}. However, their effectiveness can be hindered by two key limitations. First, large training datasets often contain substantial redundancy and noise, which inflates computational requirements and reduces training efficiency. Second, BERT’s fixed vocabulary, derived from general-domain corpora, lacks the flexibility to capture the nuanced, domain-specific language characteristic of hate speech—particularly terms involving new slurs, internet jargon, or masked profanity.

To address these challenges, we propose a data- and vocabulary-efficient fine-tuning approach for hate speech detection using BERT. Our method comprises two components: (1) a TF–IDF-based sample selection process that retains only the most informative 75\% of training data, reducing redundancy while preserving performance, and (2) a vocabulary augmentation mechanism that extends BERT’s tokenizer with hate-related slang and lexical variants absent from its original vocabulary. This dual strategy enables more efficient and context-aware learning from limited data.

We evaluate our approach on the Davidson~\cite{davidson} hate speech dataset, comparing a baseline BERT model trained on the full dataset with default vocabulary against our optimized model trained on the filtered dataset and extended lexicon. Despite the reduced dataset size, our method matches or outperforms the baseline across all major evaluation metrics while significantly reducing training time and computational cost. These findings suggest that targeted data selection and vocabulary enhancement can effectively improve the scalability and robustness of hate speech detection systems.

The main contributions of the paper can be highlighted as follows.
\begin{itemize}
    \item Introduce a lightweight data filtering technique that retains only the top 75\% most informative tweets from the training set, significantly reducing dataset size while preserving classification performance.
    \item Extend BERT’s tokenizer with hate-related slang, obfuscated terms, and variant spellings to improve coverage of abusive language not captured in the original vocabulary.
    \item Demonstrate that the proposed strategy achieves comparable or better performance across standard evaluation metrics using the Davidson et al. hate speech dataset, while reducing training time and computational overhead.
\end{itemize}


\section{Related Works}
Ali et al. (2025)~\cite{ali2025} proposed an embedding-based method to expand hate speech lexicons in real-time, enabling the detection of emerging and obfuscated slurs. They developed a hybrid model that integrates BERT with lexicon-based techniques to effectively recognize toxic language containing intentional spelling variations designed to evade detection. Their model achieved up to 95\% accuracy across several benchmark datasets, including Davidson et al.~\cite{davidson}, Founta et al.~\cite{founta2018}, Implicit Hate Corpus~\cite{implicit2021}, HateCheck~\cite{HateCheck2021}, ToxicSpan~\cite{ToxicSpan2022}, and ToxicGen~\cite{ToxicGen2022}. Additionally, they curated a secondary dataset from Twitter’s 1\% Public Streaming API (Jan 2021–Dec 2022) using random sampling. To discover obfuscated toxic terms, they employed a graph-based similarity approach for lexical expansion.

Caselli et al. (2021) introduced HateBERT~\cite{caselli-etal-2021-hatebert}, a variant of BERT fine-tuned on abusive language. They trained the model on the RAL-E dataset, which contains English Reddit comments from communities banned for offensive or hateful content. Experimental results demonstrated that HateBERT consistently outperforms standard BERT across multiple hate speech detection benchmarks.

Wang et al.~\cite{Wang2021} focused on improving the efficiency of BERT for processing long texts by applying feature selection techniques to reduce input length. They experimented with six approaches: four traditional methods—TF-IDF~\cite{TFIDF2003}, TextRank~\cite{TextRank2004}, Mutual Information (MI)~\cite{MI2004}, and Chi-squared Statistic (CHI)~\cite{chi2_2001}—as well as two truncation strategies: tail-only (preserving the last L tokens) and head+tail (preserving the first and last L/2 tokens). These methods proved effective in maintaining model performance while significantly lowering computational costs, particularly for BERT input lengths of 128, 256, or 512 tokens.

\section{Proposed Method}
\begin{figure}[htbp]
\vspace{-1cm}
    \centering
    \includegraphics[width=0.8\textwidth, trim={0 0 0 3cm},clip]{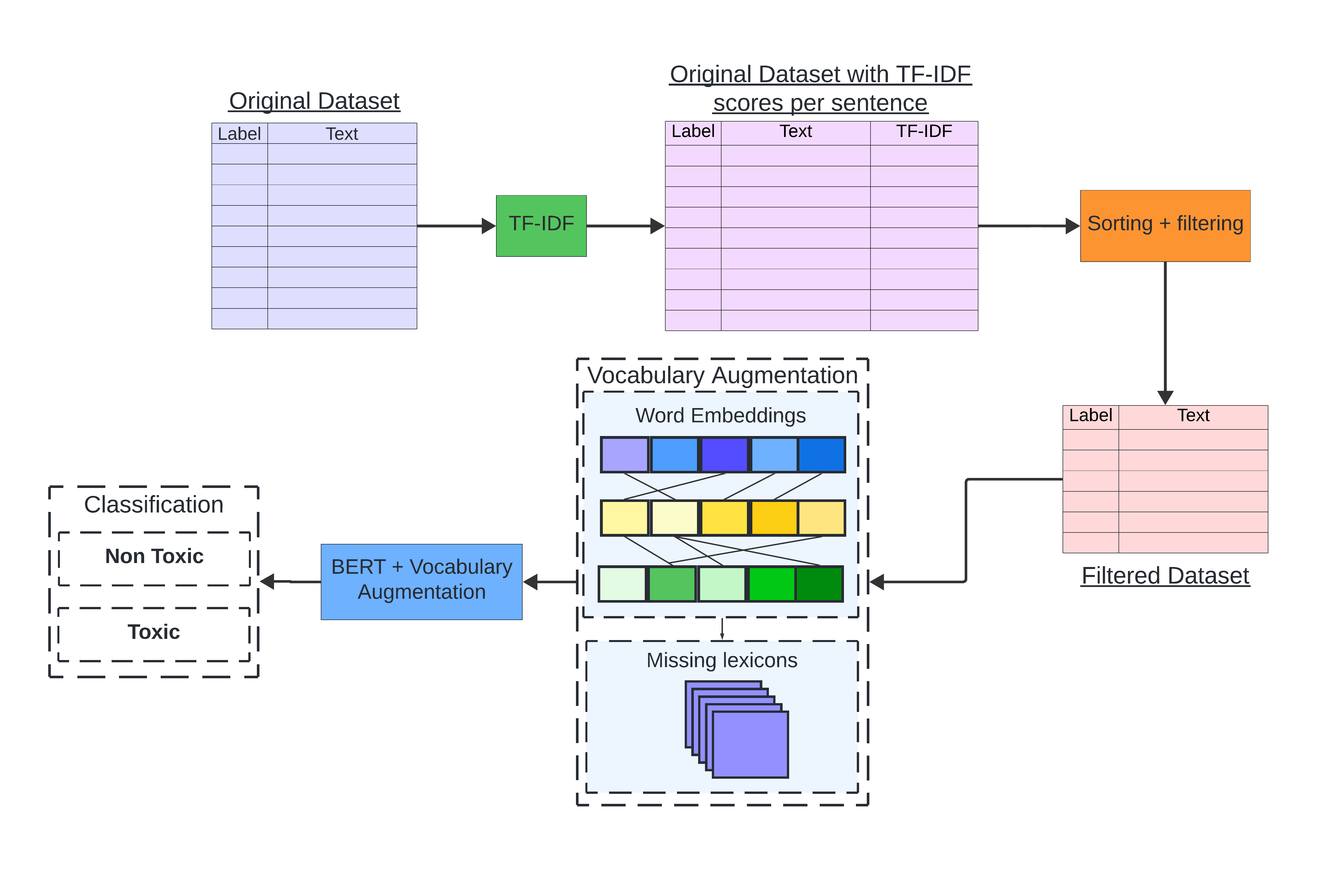} 
    \vspace{-1cm}
    \caption{Architecture of our method}
    \label{fig:arch} 
    \vspace{-0.7cm}
\end{figure}
Our method is depicted through Fig.~\ref{fig:arch}. One can infer from the above architecture that our method broadly uses two major blocks: Data filtering using TF-IDF scores, and augmentation of missing lexicons with BERT's vocabulary. 
\subsection{TF-IDF}
TF-IDF (Term Frequency–Inverse Document Frequency) is a widely used statistical measure in text mining and information retrieval. It evaluates how important a word is to a specific document relative to a collection or corpus of documents. Term Frequency (TF) measures how frequently a term $t$ appears in a document $d$. It is computed as
\begin{eqnarray}
 \text{TF}(t, d) = \frac{\text{Number of times term } t \text{ appears in } d}{\text{Total number of terms in } d}  .
\end{eqnarray}
On the other hand, Inverse Document Frequency (IDF) quantifies the rarity of a term across all documents in the corpus. It is calculated as
\begin{eqnarray}
\text{IDF}(t) = \log\left(\frac{N}{1 + \text{DF}(t)}\right),
\end{eqnarray}
where $N$ is the total number of documents, $\text{DF}(t)$ is the number of documents in which the term $t$ appears. The constant 1 is added in the denominator to avoid division by zero. The final TF-IDF score for a term is obtained by multiplying its TF and IDF values.
\begin{eqnarray}
    \text{TF-IDF}(t, d) = \text{TF}(t, d) \times \text{IDF}(t)
\end{eqnarray}
This score is high for terms that are frequent in a document but rare across the corpus, making them valuable keywords.





\subsection{TF–IDF-Based Data Filtering}
We use a fixed 80–20 train–test split on the Davidson et al. dataset, which remains constant across all experiments. In the baseline setup, we fine-tune the standard BERT model on the full 80\% training set and evaluate it on the unchanged test set.

To investigate data reduction, we apply a classical TF–IDF–based filtering strategy that retains tweets with the most distinctive linguistic features related to hate or offensive speech, while discarding those that are generic or redundant. Specifically, we compute TF–IDF scores for all words in the dataset. For each tweet, we calculate an aggregate TF–IDF score by summing the individual word scores. We then rank the tweets based on these scores and construct multiple reduced training subsets by retaining the top 80\%, 75\%, 70\%, and progressively fewer of the most informative tweets. 
This approach allows us to systematically evaluate the effects of training data reduction and vocabulary augmentation—both independently and in combination—while keeping the test set and model configuration unchanged.



\subsection{BERT}
BERT (Bidirectional Encoder Representations from Transformers)~\cite{BERT} is a deep learning model that learns contextualized word embeddings using a transformer-based architecture. It is designed to capture bidirectional dependencies by applying self-attention over entire input sequences. 
Given an input sequence of tokens $ X = \{x_1, x_2, \ldots, x_n\} $, each token \( x_i \) is first mapped to an embedding vector \( e_i \in \mathbb{R}^d \), which is a sum of the token, positional, and segment embeddings:
\begin{eqnarray}
e_i = \text{TokenEmbed}(x_i) + \text{PositionEmbed}(i) + \text{SegmentEmbed}(s_i)    
\end{eqnarray}
The resulting embeddings are passed through a stack of \( L \) transformer encoder layers. Each encoder layer uses a multi-head self-attention mechanism~\cite{Transformer}, computed as
\begin{eqnarray}
 \text{Attention}(Q, K, V) = \text{softmax}\left(\frac{QK^\top}{\sqrt{d_k}}\right)V   
\end{eqnarray}
where the queries \( Q \), keys \( K \), and values \( V \) are obtained by linear transformations of the input
$ Q = XW^Q,\quad K = XW^K,\quad V = XW^V  $
and \( W^Q, W^K, W^V \) are learnable weight matrices. BERT is trained using two self-supervised learning objectives
\begin{itemize}
  \item \textbf{Masked Language Modeling (MLM):} Random tokens in the input sequence are replaced with a special [MASK] token, and the model learns to predict the original token from the surrounding context. For a masked token \( x_m \), the probability distribution over the vocabulary is given by $    P(x_m | X_{\setminus m}) = \text{softmax}(W_o h_m)$, 
  where \( h_m \) is the hidden state at the masked position, and \( W_o \) is the output projection matrix.
  \item \textbf{Next Sentence Prediction (NSP):} The model is given a pair of sentences and trained to predict whether the second sentence logically follows the first. This is implemented as a binary classification task using the [CLS] token representation.
\end{itemize}
The output of BERT is a sequence of contextualized embeddings $ \{h_1, h_2, \ldots, h_n\}$, where each \( h_i \in \mathbb{R}^d \) represents a token enriched with contextual information from the entire sentence. These embeddings can be fine-tuned for downstream tasks such as classification, question answering, or named entity recognition.

\subsection{BERT Vocabulary Augmentation}
Reducing the size of the training dataset can limit the linguistic diversity available to the model. To address this, we introduce a complementary strategy: augmenting BERT’s vocabulary with domain-specific terms related to hate speech. The base uncased BERT model uses a fixed WordPiece vocabulary of approximately 30,000 tokens, learned from general-purpose English corpora. As a result, many slurs, slang, and creatively spelled abusive terms commonly found in hateful content are poorly represented—particularly when training data is limited.
To overcome this limitation, we curate a list of such terms from both the dataset and external hate speech lexicons, and add them to BERT’s vocabulary. Using the HuggingFace toolkit, we load the pre-trained BERT model and tokenizer, then append the new tokens. This process only alters the input embedding layer without retraining the full model, making it a lightweight yet effective form of domain adaptation. The number of added tokens remains small—just a few dozen—compared to the original 30,000-token vocabulary.
By infusing BERT with a hate-aware vocabulary, we ensure that even a reduced training set provides the model with exposure to key hateful expressions, thereby enhancing its detection capability.

\section{Experimental Setup and Results}
\subsection{Dataset and Splits}
We conduct experiments on the Davidson et al. (2017) hate speech dataset \cite{davidson}, which contains 24,783 tweets labeled into three categories: hate speech (about 5.77\% of tweets), offensive language (77.4\%), and neither (16.8\%). For simplicity, we treat tweets containing hate speech and offensive language as a single class, and tweets containing neither as the other class.

We use the dataset in two forms: the full dataset (for the baseline) and the filtered subset. To ensure fair evaluation, we fix a 20\% test split (approximately 5k tweets), which remains identical across all experiments. This test set is stratified to include a representative mix of the two classes. The remaining 80\% of the data (~19k tweets) serves as the training pool for the baseline model. For our filtered approach, we apply TF–IDF filtering to this training pool (without modifying the test set) to create a condensed training set of roughly 14k tweets. The test set always remains the same, enabling a direct, apples-to-apples comparison of model performance.

\subsection{Results}
Here, we explore the impact of TF-IDF-based dataset filtering and vocabulary expansion on the performance of a BERT-based hate speech classification model. By systematically filtering the training data to retain the most informative samples and enhancing the tokenizer with the explicit inclusion of frequently occurring offensive terms, we aim to improve the model’s generalization and sensitivity to critical linguistic cues. 
Fig.~\ref{fig:bar} presents a comparative analysis of key evaluation metrics across different filtering thresholds, highlighting the balance between dataset reduction and representational power.
\begin{figure}[H]
\vspace{-0.8cm}
    \centering
    \includegraphics[width=0.7\textwidth]{comparisonChart1.png} 
    \vspace{-0.5cm}
    \caption{Comparison of evaluation metrics (Accuracy, Precision, Recall, and F1 Score) for the vanilla model and various TF-IDF filtered variants (from 80\% to 50\%).}
    \vspace{-0.7cm}
    \label{fig:bar} 
\end{figure}
The plot illustrates that the vanilla BERT model maintains consistent performance across all evaluation metrics, establishing a strong baseline. In contrast, the TF-IDF-based variants reduce the training data by retaining only the top-\textit{k} most informative samples, while simultaneously expanding BERT’s vocabulary with domain-specific offensive terms. This twofold strategy mitigates noise and enhances tokenization for linguistically obfuscated or abusive expressions. The results indicate that moderate levels of filtering ($\sim$ 75\%–70\%) either preserve or slightly improve classification performance, suggesting that a more focused dataset, when combined with targeted vocabulary augmentation, enhances model robustness. However, more filtering (60\%–50\%) leads to a noticeable performance drop, implying that excessive data reduction compromises essential linguistic context. These findings underscore the importance of striking a balance between dataset selectivity and the preservation of linguistic diversity.



\begin{table}[!htbp]
\vspace{-0.5cm}
\centering
\caption{Detailed Class-wise Evaluation}
\vspace{-0.3cm}
\label{tab:classwise_eval_65}
\resizebox{\textwidth}{!}{
\begin{tabular}{|l|c|c|c|c|c|c|c|c|c|}
\hline
\textbf{Metric} & \textbf{Original} & \textbf{TF-IDF } & \textbf{TF-IDF } & \textbf{TF-IDF } & \textbf{TF-IDF} & \textbf{TF-IDF} & \textbf{TF-IDF} & \textbf{HurtBERT}& \textbf{BERT+}\\
 & \textbf{Davidson} & \textbf{top 80\%} & \textbf{top 75\%} & \textbf{top 70\%} & \textbf{top 65\%} & \textbf{top 60\%} & \textbf{top 50\%} & \cite{HurtBERT2020} & {\bf Deep Model}\cite{deepmodel2021}\\
\hline
Precision (Class 0) & 0.8812 & 0.8679 & \textbf{0.9044} & 0.8757 & 0.8692 & 0.8696 & 0.8905 & - & - \\
Recall (Class 0)    & 0.8992 & \textbf{0.9304} & 0.8968 & 0.9136 & 0.9256 & 0.9208 & 0.8595 & - & - \\
F1-score (Class 0)  & 0.8901 & 0.8980 & \textbf{0.9005} & 0.8942 & 0.8965 & 0.8980 & 0.8748 & - & 0.89 \\
\hline
Precision (Class 1) & 0.9795 & \textbf{0.9857} & 0.9792 & 0.9824 & 0.9848  & 0.9838 & 0.9718 & - & 0.96 \\
Recall (Class 1)    & 0.9755 & 0.9714 & \textbf{0.9808} & 0.9738 & 0.9719 & 0.9714 & 0.9787 & - & 0.96 \\
F1-score (Class 1)  & 0.9775 & 0.9785 & \textbf{0.9800} & 0.9781 & 0.9783 & 0.9785 & 0.9752 & - & {\bf 0.98} \\
\hline
Macro Avg F1  & 0.9338 & 0.9383 &{\bf  0.9403} & 0.9362 & 0.9374 & 0.9362 & 0.9250 & 0.908 & - \\
Weighted Avg F1 & 0.9628 & 0.9650 & {\bf 0.9667} & 0.9640 & 0.9645 & 0.9639 & 0.9584 & - & 0.962 \\
\hline
\end{tabular}
}
\vspace{-0.5cm}
\end{table}
\begin{table}[!htbp]
\vspace{-0.8cm}
\centering
\caption{Comparison of BERT Performance on Original vs. TF-IDF Filtered Davidson Dataset}
\vspace{-0.3cm}
\label{tab:training_time_accuracy}
\begin{tabular}{|l|c|c|}
\hline
\textbf{Configuration} & \textbf{Total Time (s)} & \textbf{Accuracy} \\
\hline
Original Davidson (80-20 split) & 1420.13 & 0.9627 \\
TF-IDF Top 80\% (training data) & \textbf{1086.69} & \textbf{0.9645} \\
TF-IDF Top 75\% (training data) & \textbf{1039.14} & \textbf{0.9667} \\
TF-IDF Top 70\% (training data) & \textbf{973.68} & \textbf{0.9637} \\
TF-IDF Top 65\% (training data) & \textbf{919.93} & \textbf{0.9641} \\
TF-IDF Top 60\% (training data) & \textbf{870.61} & \textbf{0.9635} \\
TF-IDF Top 50\% (training data) & \textbf{743.96} & 0.9586 \\
\hline
\end{tabular}
\vspace{-0.7cm}
\end{table}
 Tables 1 and 2 demonstrate the effectiveness of the proposed approach. Table 1 provides a detailed breakdown of precision, recall, and F1 scores across various levels of training data filtering. Notably, at the 75\% and 70\% thresholds, performance remains stable or slightly improves across both classes, indicating that eliminating low-information tweets enables the model to better focus on informative linguistic patterns. This suggests that moderate data reduction can enhance generalization. However, performance begins to decline below the 65\% level—particularly in Class 0 precision—highlighting the adverse impact of reduced linguistic diversity. Despite this, even the 65\% filtered dataset outperforms many existing methods in most evaluation metrics. Table 2 further shows that TF-IDF-based filtering significantly reduces training time—by nearly 50\% at the 50\% threshold—while maintaining accuracy close to or exceeding the baseline, underscoring the computational efficiency of the proposed strategy.
 \begin{table}[!htbp]
\centering
\vspace{-0.5cm}
\caption{Comparison of BERT Performance on Original vs. TF-IDF Filtered Davidson Dataset without vocabulary augmentation (epochs = 3, batch size = 8)}
\vspace{-0.3cm}
\label{tab:voc}
\begin{tabular}{|l|c|c|}
\hline
\textbf{Configuration} & \textbf{Total Time (s)} & \textbf{Accuracy} \\
\hline
Original Davidson (80-20 split) & 1420.13 & 0.9627 \\
TF-IDF Top 80\% (training data) & \textbf{1066} & 0.9613 \\
TF-IDF Top 70\% (training data) & \textbf{955.10} & \textbf{0.9631} \\
TF-IDF Top 60\% (training data) & \textbf{854.04} & 0.958 \\
TF-IDF Top 50\% (training data) & \textbf{729} & 0.955 \\
\hline
\end{tabular}
\vspace{-0.8cm}
\end{table}
We further assess the impact of vocabulary augmentation in Table~\ref{tab:voc}. The results show that, across all filtering levels, the absence of augmentation leads to a consistent drop in accuracy, underscoring the importance of incorporating domain-specific vocabulary. These findings highlight the interplay between dataset quality and representational capacity. While TF-IDF filtering effectively removes noisy or redundant samples, directing learning toward more informative examples, vocabulary augmentation enhances the model's ability to recognize adversarial slang and obfuscated hate terms that standard BERT tokenizers often overlook. Together, these techniques enable efficient learning from fewer examples. However, excessive filtering may limit the model’s exposure to critical contextual information, reducing its ability to detect implicit forms of hate. Therefore, a balanced strategy—combining moderate filtering with vocabulary augmentation—offers the optimal trade-off between efficiency and model robustness.

\section{Conclusion}
\vspace{-0.3cm}
This paper presents an efficient and scalable approach for hate speech detection that balances model performance with reduced computational overhead. By leveraging TF–IDF-based filtering, we identify and retain the most informative training examples, substantially lowering data volume without sacrificing classification accuracy. Furthermore, we address the lexical limitations of BERT by augmenting its vocabulary with domain-specific slang and obfuscated terms prevalent in abusive language. Results confirm that our combined strategy not only maintains competitive performance but also enhances training efficiency. These findings underscore the viability of lightweight adaptations for real-world deployment in dynamic and evolving social media environments.
\vspace{-0.5cm}
\bibliographystyle{plain}
\bibliography{Refer}

\end{document}